\documentclass[runningheads]{llncs}
\usepackage{graphicx}
\usepackage{amssymb}
\usepackage{soul}
\usepackage{authblk}
\usepackage{amsmath}
\usepackage[misc]{ifsym}
\usepackage{orcidlink}
\definecolor{lightblue}{RGB}{173,216,230}
\sethlcolor{lightblue}
\usepackage{algorithm}
\usepackage{algorithmicx}
\usepackage{algpseudocode}
\usepackage[ruled,vlined,algo2e]{algorithm2e}

\usepackage{booktabs}
\usepackage{multirow}

\newcommand{\mythanks}{These authors contributed equally to this work.}
\begin{document}
\title{Stage-wise Adaptive Label Distribution for Facial Age Estimation}
%
%

\author{Bo Wu$^{1,}$\thanks{\mythanks} \and
Zhiqi Ai$^{1,}$\textsuperscript{$\star$} \and
Jun Jiang$^{1,}$\textsuperscript{$\star$} \and
Congcong Zhu$^2$ \and
Shugong Xu\textsuperscript{(\Letter)}$^{,3}$
}

\institute{$^1$Shanghai University, Shanghai, China\\
$^2$University of Science and Technology of China (USTC), China\\
$^3$Xi’an Jiaotong Liverpool University, China \\
\email{aizhiqi-work@shu.edu.cn, shugong.xu@xjtlu.edu.cn}\\ }

%
%

\maketitle              
\begin{abstract}

Label ambiguity poses a significant challenge in age estimation tasks. Most existing methods address this issue by modeling correlations between adjacent age groups through label distribution learning. However, they often overlook the varying degrees of ambiguity present across different age stages. In this paper, we propose a \textbf{\underline{S}}tage-wise \textbf{\underline{A}}daptive \textbf{\underline{L}}abel \textbf{\underline{D}}istribution \textbf{\underline{L}}earning (SA-LDL) algorithm, which leverages the observation—revealed through our analysis of embedding similarities between an anchor and all other ages—that label ambiguity exhibits clear stage-wise patterns. By jointly employing stage-wise adaptive variance modeling and weighted loss function, SA-LDL effectively captures the complex and structured nature of label ambiguity, leading to more accurate and robust age estimation. Extensive experiments demonstrate that SA-LDL achieves competitive performance, with MAE of 1.74 and 2.15 on the MORPH-II and FG-NET datasets.

\end{abstract}
%
%
%





\section{Introduction}

Facial age estimation is a common task in face analysis with applications in security, marketing, and healthcare. Although deep learning methods \cite{dex,method2,TIAN2021158} have advanced this field, label ambiguity remains a major challenge because faces of adjacent ages often look visually similar \cite{BAO202386,c3ae,ol}.

Label distribution learning (LDL) \cite{dldl,geng2013facial,24,25,26} mitigates this issue by considering neighboring age labels when assigning labels. LDL assumes smooth transitions between adjacent ages in the feature space, improving accuracy and robustness. However, LDL treats all samples uniformly, without adapting to individual differences.

Adaptive label distribution learning (A-LDL) methods \cite{geng,avdl,he,hou,pan} improve upon LDL by learning distributions specific to each age or sample, accounting for variability across age stages. Yet, A-LDL often requires cross-age data from the same individual to model aging trajectories, which is rarely available in public datasets. Additionally, per-sample adaptation can lead to overfitting.

As illustrated in Figure \ref{fig1}, sub-figures (a)-(f) compare label ambiguity for the same person at different life stages, showing notable differences between childhood and adolescence but little variation within each stage. Sub-figure (g) further confirms this by displaying age-specific variance distributions that remain relatively stable within certain age ranges. This observation—revealed through our analysis of embedding similarities between an anchor and all other ages—indicates that label ambiguity exhibits clear stage-wise patterns, as shown in Figure \ref{fig2}.

To leverage this insight, we propose a stage-wise adaptive label distribution learning (SA-LDL) algorithm. SA-LDL models label ambiguity across grouped age stages by incorporating stage-wise adaptive variance and weighted loss function. This design allows the model to effectively capture the complex and structured nature of label ambiguity, leading to improved accuracy and robustness in age estimation. Our contributions can be summarized as follows:

\begin{figure}[t]
\centering
\includegraphics[width=3.5in]{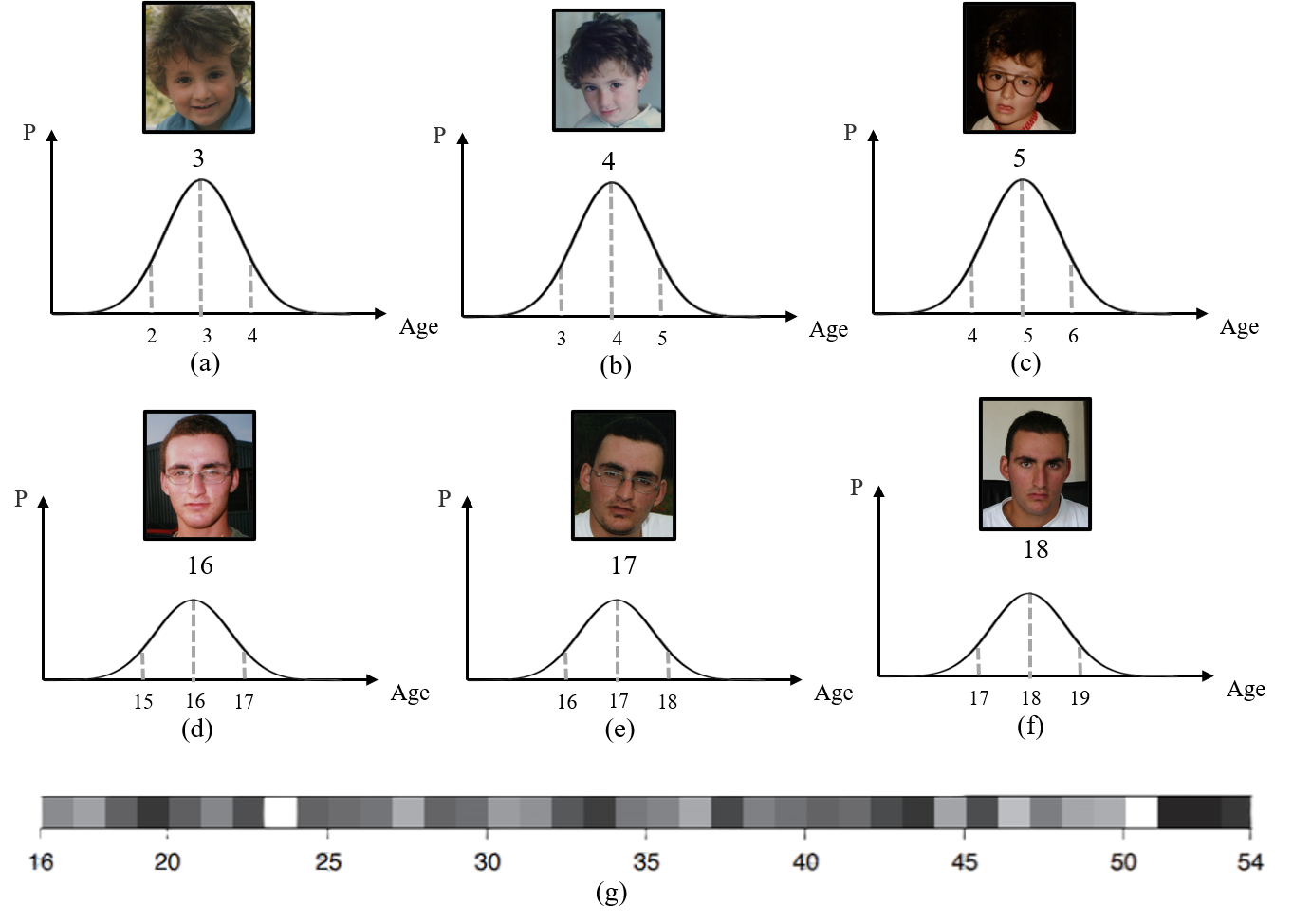}
\caption{Stage-wise label ambiguity.}
\label{fig1} 
\end{figure}  

\begin{figure}[!t]
\centering
\includegraphics[width=\textwidth]{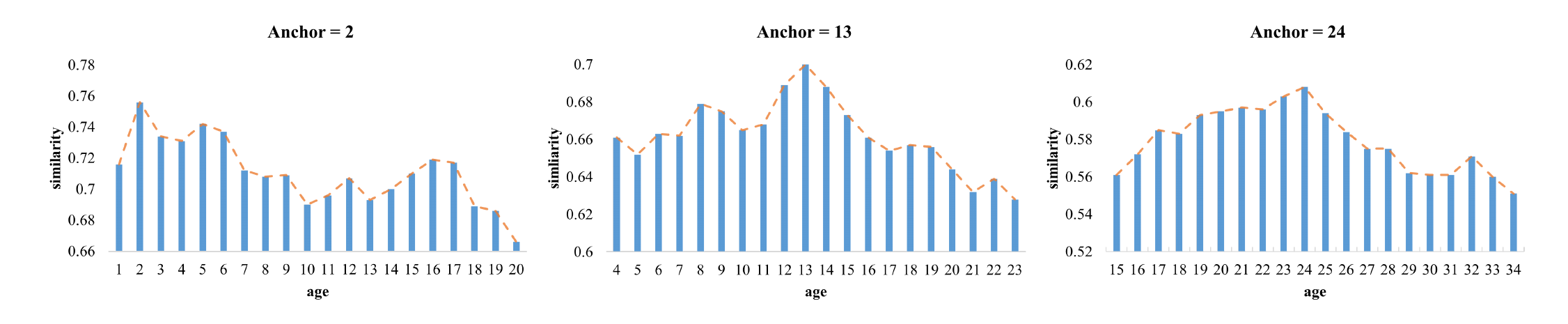}
\caption{Cosine similarity data analysis of different anchor.}
\label{fig2} 
\end{figure}

\begin{itemize}
    \item{We propose a stage-wise label distribution learning (SA-LDL) algorithm that effectively captures the variation of label ambiguity across different age stages. It achieves grouped adaptive modeling of label uncertainty over the temporal dimension, enhancing the model’s understanding of age-related features.}
    \item{We propose a stage-wise adaptive variance (SAV) learning algorithm and stage-wise adaptive weighted (SAW) loss function improving the model’s capacity to represent the complex structure of label ambiguity.}
    \item{Extensive experiments demonstrate that SA-LDL achieves competitive performance, with MAE of 1.74 and 2.15 on the MORPH-II and FG-NET datasets.}
\end{itemize}

\section{Related Work}

\subsection{Facial age estimation}
With the advancement of deep learning, various deep structures have been widely applied to facial age estimation tasks, mainly including regression, classification, and ranking methods \cite{survey2,survey1}. Classification-based methods typically treat age estimation as a classification problem, where age is viewed as an independent label, and the correlation between labels cannot be leveraged \cite{dex,regression2}. \cite{dex} proposed an algorithm combining Convolutional Neural Networks (CNN) with classification expected output, which significantly enhanced the performance of age estimation. \cite{regression2} conducted a quantitative evaluation of various classifiers in automatic age estimation tasks, testing classifiers based on quadratic functions, shortest distance, and artificial neural networks. Regression-based methods generally use Euclidean loss to regress the true age value, penalizing the difference between estimated and true values \cite{18,19,20}. \cite{18} introduced ordinal-aware manifold analysis to find low-dimensional subspaces, then applied a multivariate linear regression model to constrain the relationship between low-dimensional features and the true age values. \cite{19} and \cite{20} used CNN models to extract features and employed squared loss for age regression estimation. Ranking-based methods convert the ordinal target into multiple binary classification sub-tasks using neural networks \cite{method3,22,coral}. \cite{method3} leveraged ordinal information by learning a network with multiple binary outputs, while \cite{22} aggregated the outputs of multiple binary CNNs for age estimation. \cite{coral} addressed the inconsistency issue between binary classifiers in ranking methods and provided theoretical support for the monotonicity and consistency of ranking confidence scores.


\subsection{Label distribution learning}

Label distribution learning (LDL) methods provide flexibility in handling the uncertainty and ambiguity in facial age estimation by learning a probability distribution over predefined age categories rather than assigning a single label, thus better capturing the diversity of age estimation \cite{dldl,geng2013facial,24,25,26}. LDL is an effective solution to the label ambiguity problem. Early research focused primarily on fixed-form label distributions, assuming a Gaussian distribution for age labels with the real age as the mean and predefined variance, using soft labels from the distribution to replace hard labels, aiming to minimize the gap between the learned and fixed distributions.

\cite{geng2013facial} introduced label distribution into age estimation tasks, treating facial images as instances associated with label distributions, and defined the label distribution by assigning Gaussian or triangular distributions to these instances. \cite{24} were the first to explain the label ambiguity problem and used CNNs to minimize the relative entropy (Kullback-Leibler, KL) between two Gaussian distributions for label distribution learning. Furthermore, \cite{25} studied the relationship between label distribution learning and ranking-based methods, theoretically proving that ranking methods are a special case of label distribution learning algorithms, and experimentally demonstrated the effectiveness of label distribution. \cite{26} proposed an end-to-end deep regression forest model, combining random forests with neural networks to model general form label distributions using the decision tree’s capabilities. While these methods alleviated the label ambiguity issue to some extent, they are limited by their dependence on fixed-form label distribution assumptions, which restricts the expressiveness of the learned features.


\subsection{Adaptive label distribution learning}

Adaptive label distribution learning (A-LDL) is a method that aims to address the issue of label ambiguity and improve representations in age estimation tasks \cite{geng,avdl,he,hou,pan}. \cite{geng} developed a technique that iteratively updates the variance for each age, allowing the generation of an adaptive label distribution. \cite{avdl} argued that variance not only depends on the correlation between adjacent ages but also varies based on the age and identity of the sample. To address this, they proposed a data-driven meta-learning optimization framework that models the variance for specific samples. \cite{he} presented a data-driven label distribution algorithm that does not rely on any prior information about the learning form. Instead, it generates an age label distribution by combining the input image's label with similar samples in its context, leveraging the similarities between samples. Furthermore, \cite{hou} observed that adapting the label distribution requires more training data. To overcome this challenge, they introduced semi-supervised adaptive label distribution learning, where unlabeled data is utilized to enhance the adaptation process. Lastly, \cite{pan} proposed the mean-variance loss, a novel loss function designed to optimize the label distribution. The mean loss aims to minimize the difference between the estimated distribution's mean and the true value of the age label, and the variance loss, on the other hand, controls the sharpness of the estimated distribution. These ensure that the obtained distribution is accurate and well-defined.

\section{Methodology}

\subsection{Preliminary}

Label distribution learning (LDL) convert age into a Gaussian distribution with a mean of real age $ y $ and a standard deviation of $ \sigma $. Use $ d_k(y,\sigma) $ to represent the $k$-dimensional label distribution vector, which is the probability of the true age being $ k $ years old, where $k\in [0,100]$.
\begin{equation}
\label{dk}
d_k(y,\sigma) = \frac{1}{\sqrt{2\pi}\sigma}exp(-\frac{(k-y)^2}{2\sigma^2}).
\end{equation}

When predicting, the last 101-dimensional classification vector $ z(X,\theta)$ is transformed into a probability distribution $\hat{d}_k$, and its $k$-dimensional probability:
\begin{equation}
\label{dkhat}
\hat{d}_k(X,\theta) = \frac{exp(z_k(X,\theta ))}{{\textstyle \sum_{n}^{}exp(z_n(X,\theta))}}.
\end{equation}
where $X$ is the input image and $\theta$ is the model parameters. 

The constraint between the predicted probability distribution generated by the model and the true label probability distribution is usually measured using KL divergence to measure the difference between the two probability distributions:
\begin{equation}
\label{KLLoss1}
L_{KL}(X,y,\theta,\sigma) = \sum_{k}d_k(y,\sigma)ln{\frac{d_{k}(y,\sigma)}{\hat{d}_{k}(X,\theta)}}.
\end{equation}

\subsection{Stage-wise label distribution learning (SA-LDL)}

\begin{figure}[h]
\centering
\includegraphics[width=\textwidth]{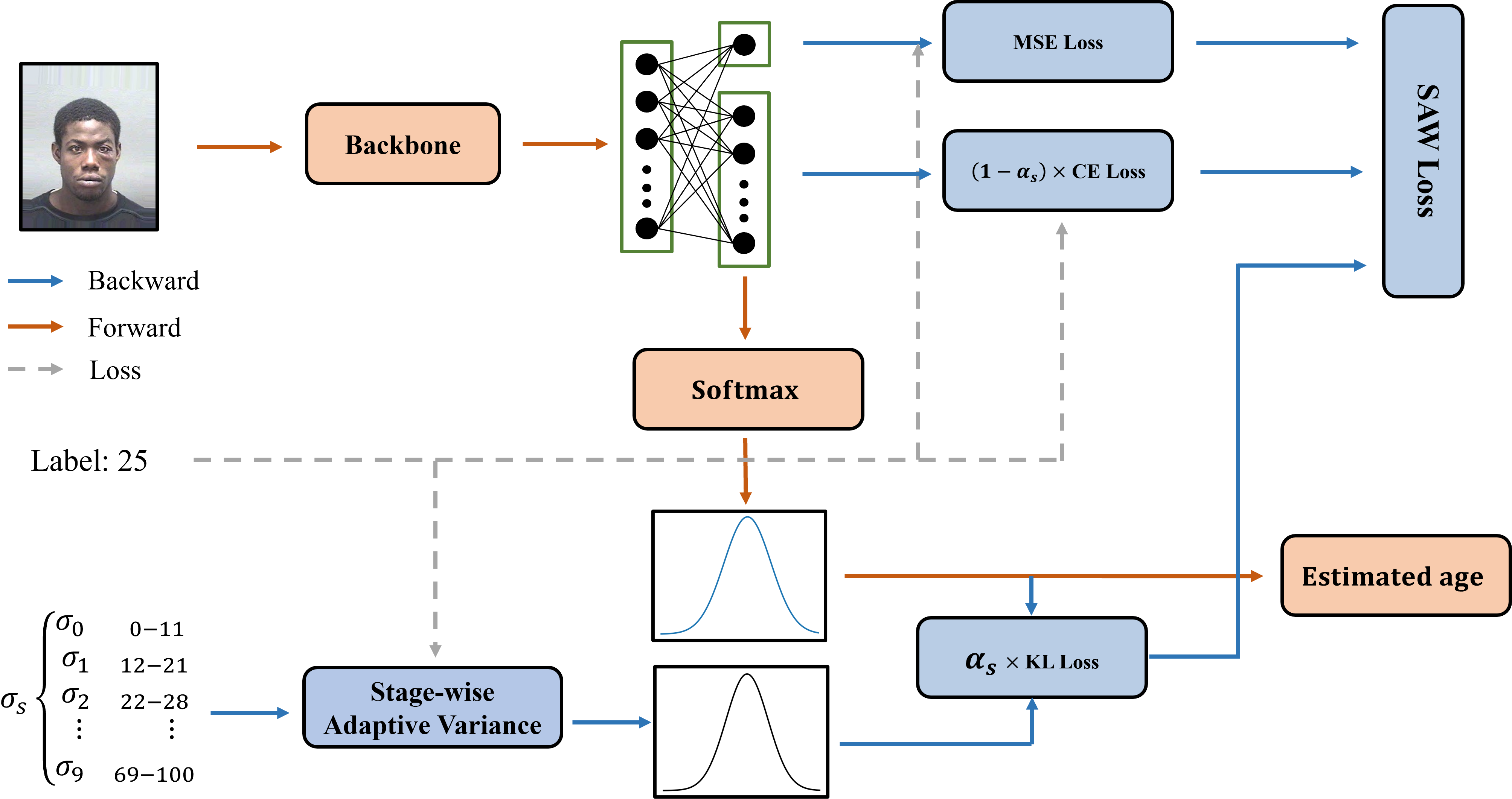}
\caption{The pipeline of the proposed SA-LDL, consisting of three components: the feature extraction backbone, the stage-wise adaptive variance learning algorithm (SAV), and the stage-wise adaptive weight loss function (SAW).}
\label{fig_3}
\end{figure}

Stage-wise label ambiguity is a ubiquitous phenomenon observed across different individuals as in Figure \ref{fig2}. We presents the stage-wise adaptive label distribution learning (SA-LDL), as shown in Figure \ref{fig_3}. The algorithm consists of three main components: the feature extraction backbone (such as EfficientNetV2 \cite{tan2021efficientnetv2}), the stage-wise adaptive variance learning algorithm (SAV), and stage-wise adaptive weight loss function (SAW).

Traditional age segmentation methods divide age into ten-year intervals. Although this approach is simple to implement, it fails to provide an optimal data distribution for each stage, which affects the effectiveness of model training. To address this, we employ the K-means clustering algorithm \cite{kmeans} to cluster labels from multiple datasets \cite{MORPH,fgnet,utk,cacd}. By clustering, we can group data based on similar facial features, allowing us to identify common patterns and variations in facial changes.

\subsection{Stage-wise adaptive variance learning (SAV)}
Current adaptive label distribution algorithms still face issues with adaptive variance. First, the learning of adaptive variance may fluctuate and is synchronized with the update of model parameters, making it difficult to precisely consider the impact for each age. Secondly, overfitting is a concern, especially when datasets are mismatched. Lastly, the root cause of label ambiguity lies in the different rates at which facial features change across ages.

To address these challenges, we propose the stage-wise adaptive variance learning (SAV) algorithm. The age range of 0-100 years is divided into 10 stages, with each stage learning a separate variance. This approach expands the variance range, reduces fluctuations when learning variance for each age, and helps to achieve more stable variances. Additionally, by repeating this learning process across multiple datasets, we can obtain a variance range applicable to different datasets.

The SAV (see Algorithm \ref{alg}) initializes a 10-dimensional standard deviation vector to 0, based on the stage-wise nature of label ambiguity. The model is then trained with KL Loss \eqref{KLLoss1} and validated, with the standard deviation updated through adaptive learning. The update criterion is that if the L1 loss, $L_1 = |y_{pred} - y_{val}|$, on the validation set decreases compared to the previous iteration, the model parameters and standard deviation are updated.

\begin{algorithm}
    \caption{Stage-wise adaptive variance learning (SAV).}
    \label{alg}
    \setcounter{AlgoLine}{0}
    \scriptsize
    \LinesNumbered
    \KwIn{train dataset $S_{tr}=(X_{tr},y_{tr})$, val dataset $S_{val}=(X_{val},y_{val})$, initial model $\theta_{0}$, initial standard deviation $\sigma_{0}$}
    \KwOut{model $\theta^{t}$, stage standard deviation $\sigma^{t}$}
    
    Initial minimize loss $ \min L_1 = \infty $
    \;
    \;
    
    \For{$ i=0,1... epoch $}{
        Train model $\theta_{i}$ with $L_{KL}$ \eqref{KLLoss1} with $S_{tr}$
        
        Calculate $L_1$ and $L_{KL}$ \eqref{KLLoss1} with $S_{val}$
        
        {
            \If{$L_1 < \min L_1$}{
              Update model $\theta_{\hat{i}}$ and standard deviation  $\sigma_{\hat{i}}$ 

                $ \min L_1 = L_1$
            }
        }
    }

    \textbf{Output} \;
\end{algorithm}

\subsection{Stage-wise adaptive weighted loss function (SAW)}
Label ambiguity is reflected not only in the variance but also in the loss function. The degree of label ambiguity varies across different age groups. For age groups with lighter label ambiguity (e.g., 0-11 years old), classification loss has a more significant impact than distribution loss. However, for age groups with heavier label ambiguity (e.g., 22-28 years old), distribution loss is more reliable. This is because classification loss struggles to distinguish similar features, while distribution loss can differentiate based on the distribution form. Therefore, the loss function should be designed in stages. Age estimation should be viewed as a coupled problem, rather than a simple classification, regression, or distribution problem, with different stages treated as distinct subproblems.

If age estimation is treated as a classification problem, cross-entropy loss $L_{ce}$ is commonly applied for constraints.
\begin{equation}
\label{CELoss}
L_{ce} = - \frac{1}{N} \sum_{i=1}^{N} \sum_{c=1}^{M} y_{i,c} log{(\hat{y}_{i,c})}.
\end{equation}

If age estimation is defined as a regression problem, mean squared error loss $L_{mse}$ is commonly applied.

\begin{equation}
\label{MSELoss}
L_{mse} = \frac{1}{N} \sum_{i=1}^{N} {(\hat{y}_{i}-y_{i})}^{2}.
\end{equation}

If age estimation is treated as a label distribution problem, the relative entropy (Kullback-Leibler) loss $L_{kl}$ is commonly applied. 
\begin{equation}
\label{KLLoss2}
L_{kl} = \sum_{i=1}^{N} \sum_{c=1}^{M} y_{i,c} log{\frac{y_{i,c}}{\hat{y}_{i,c}}}.
\end{equation}

In this context, classification and distribution losses generally dominate, while regression loss acts as a supplementary term. To account for the stage-wise uncertainty of labels, a stage-wise adaptive weight (SAW) method is used to learn the weights for classification and distribution at each stage. Additionally, regression loss is applied with fixed weights. The combination of these learned losses is referred to as the stage-wise adaptive weight loss function $L_{SAW}$.

\begin{equation}
\label{SAWLoss}
L_{SAW} = \alpha \times L_{kl} + (1 - \alpha) \times L_{ce} + 0.01 \times L_{mse}
\end{equation}

\section{Experiments}
\subsection{Datasets}
We evaluate the proposed SA-LDL method on four representative datasets: MORPH-II \cite{MORPH}, FG-NET \cite{fgnet}, UTK \cite{utk}, and CACD \cite{cacd}. These datasets vary in age range, scale, and acquisition settings, allowing for a comprehensive assessment of the algorithm’s generalization ability.

MORPH-II \cite{MORPH} is a medium-sized facial age dataset containing 55,134 images from 13,617 subjects, with age labels ranging from 16 to 77 years. The images were taken under controlled conditions. The training and testing sets contain 44,228 and 11,058 images, respectively.

FG-NET \cite{fgnet} is a small dataset with 1,002 color and grayscale images from 82 subjects, covering an age range of 0 to 69 years. The images were collected from real-life scenarios. The training set contains 799 images, and the test set contains 100 images.

UTK \cite{utk} is a wide age-span dataset with 23,707 facial images labeled from 0 to 116 years, captured in unconstrained real-world conditions. The training and testing sets include 18,965 and 2,371 images, respectively.

CACD \cite{cacd} is a large-scale cross-age celebrity dataset with 163,446 images of 2,000 celebrities, collected using search engines with names and years as keywords. The training and test sets consist of 145,275 and 10,571 images, respectively.

\subsection{Implementation Details}
We implement the proposed SA-LDL using the PyTorch with EfficientNetV2 \cite{tan2021efficientnetv2} as the backbone network. The model is optimized using SGD with a base learning rate of 1e-3, a batch size of 32, and an input resolution of 224$\times$224. All experiments are conducted on a NVIDIA RTX 3090 GPU. For evaluation, we adopt two widely used metrics in facial age estimation: Mean Absolute Error (MAE) and Cumulative Score (CS), to assess the performance across all experiments.

\section{Results}
\subsection{Main Results}

We evaluate our proposed SA-LDL method on four benchmark datasets. On the MORPH-II \cite{MORPH} dataset as shown in Table 1, SA-LDL achieves a better MAE score than the current state-of-the-art Meta-Age \cite{meta-age} (1.75 vs. 1.81), representing a 3.3\% relative improvement. For the $CS$ metric, SA-LDL (92.2) performs comparably to the advanced MWR \cite{mwr} (95.0). This performance gain is mainly attributed to SA-LDL’s comprehensive modeling of label ambiguity in a stage-wise manner, combining the strengths of variance adaptation and distribution adaptation. Representative methods such as AVDL \cite{avdl} and UC \cite{uc} employ variance and distribution adaptation respectively—UC implements distribution adaptation via a loss function. Although these approaches model label ambiguity from different perspectives, they overlook its stage-wise nature. In contrast, SA-LDL explicitly incorporates the stage-wise characteristics of label ambiguity, leveraging controllable variance adaptation alongside distribution-based constraints. This integrated approach enables SA-LDL to outperform the compared methods in overall performance.

\begin{table}[ht]
\centering
\begin{minipage}{0.48\linewidth}
\label{tab:1}
\centering
\caption{Results on MORPH-II.}
\begin{tabular}{@{}ccc@{}}
\toprule
Method       & MAE (↓)           & CS(\%) (↑)        \\ \midrule
OL \cite{ol}           & 2.22          & 93.3          \\
AVDL \cite{avdl}         & 1.94          & -             \\
DRF \cite{drf}         & 2.14          & 91.3          \\
MWR \cite{mwr}          & 2.00          & \textbf{95.0} \\
FP-Age \cite{fpage}       & 1.90          & 93.7          \\
UC \cite{uc}           & 1.86          & -             \\
Meta-Age \cite{meta-age}     & 1.81          & -             \\ \midrule
SA-LDL (Ours)& \textbf{1.75} & 92.2          \\ \bottomrule
\end{tabular}
\end{minipage}
\hfill
\begin{minipage}{0.48\linewidth}
\centering
\label{tab:2}
\caption{Results on FG-NET.}
\begin{tabular}{@{}ccc@{}}
\toprule
Method       & MAE (↓)          & CS(\%) (↑)        \\ \midrule
Dex \cite{dex}          & 3.09          & -             \\
AGEn \cite{agen}         & 2.96          & 85.0          \\
C3AE \cite{c3ae}        & 2.95          & -             \\
BridgeNet \cite{Bridgenet}   & 2.56          & 86.0          \\
AVDL \cite{avdl}        & 2.32          & -             \\
DRF \cite{drf}        & 3.85          & 80.6          \\
MWR \cite{mwr}          & 2.23          & \textbf{91.1}          \\ \midrule
SA-LDL (Ours)& \textbf{2.15} & 90.2          \\ \bottomrule
\end{tabular}
\end{minipage}
\end{table}

In the FG-NET \cite{fgnet} dataset (Table 2), experimental results show that the MAE of our proposed SA-LDL (2.15) surpasses the state-of-the-art MWR algorithm \cite{mwr} (2.23), achieving a 3.6\% improvement. Regarding the $CS$ metric, SA-LDL (90.2) performs comparably to MWR (91.1).

\begin{table}[ht]
\centering
\begin{minipage}{0.48\linewidth}
\centering
\caption{Results on UTK.}
\begin{tabular}{@{}cc@{}}
\toprule
Method       & MAE (↓)             \\ \midrule
CORAL \cite{coral}          & 5.47              \\
Gustafsson \cite{gustafsson2020energy}         & 4.65               \\
Berg \cite{berg2021deep}         & 4.55             \\
MWR \cite{mwr}         & \textbf{4.37}         \\ \midrule
SA-LDL (Ours)& \underline{4.45}      \\ \bottomrule
\end{tabular}
\end{minipage}
\hfill
\begin{minipage}{0.48\linewidth}
\centering
\caption{Results on CACD.}
\begin{tabular}{@{}cc@{}}
\toprule
Method       & MAE (↓)      \\ \midrule
Dex \cite{dex}         & 4.78             \\
DRF \cite{drf}         & 4.61              \\
MWR  \cite{mwr}        & 4.41                   \\ 
FP-Age  \cite{fpage}        & 4.33                   \\ \midrule
SA-LDL (Ours)& \textbf{4.21}          \\ \bottomrule
\end{tabular}
\end{minipage}
\end{table}

The proposed algorithm achieves comparable MAE performance on the UTK dataset \cite{utk} (SA-LDL) to the state-of-the-art method MWR \cite{mwr} (4.37) as shown in Table 3. On the CACD dataset (Table 4) \cite{cacd}, experimental results show that the proposed algorithm outperforms the current leading method FP-Age \cite{fpage} in terms of MAE (SA-LDL: 4.21 vs. FP-Age: 4.33).

It is worth noting that the datasets vary in size and scale. The proposed algorithm not only delivers competitive performance on large-scale datasets but also surpasses existing state-of-the-art methods on smaller datasets, demonstrating strong generalization ability across datasets of different scales.

\subsection{Ablation Studies}

\textbf{Stage-wise adaptive variance learning.} As shown in Table \ref{tab:5}, the ablation experiment results of the stage-wise adaptive variance learning algorithm are presented on the MORPH-II \cite{MORPH}, FG-NET \cite{fpage}, UTK \cite{utk}, and CACD \cite{cacd} datasets. In this experiment, the predefined variance is set to the empirical value of 2 \cite{avdl}. Compared to the predefined variance, the stage-wise adaptive variance fully considers the stage-wise nature of label ambiguity. By generating the variance distribution according to the severity of label ambiguity in each stage, it achieves better performance.

\begin{table}[]
\centering
\caption{Ablation study on SAV.}
\resizebox{0.75\textwidth}{!}{%
\begin{tabular}{@{}cccccc@{}}
\toprule
\multirow{2}{*}{Dataset} & \multicolumn{2}{c}{Predefined (vars=2) \cite{avdl}} &  & \multicolumn{2}{c}{SAV (Ours)} \\ \cmidrule(lr){2-3} \cmidrule(l){5-6} 
                         & MAE (↓)             & CS(\%) (↑)               &  & MAE (↓)        & CS(\%) (↑)         \\ \midrule
MORPH-II \cite{MORPH}                    & 2.21            & 90.96            &  & 2.01       & 92.30      \\
FG-NET \cite{fgnet}                  & 4.63            & 63.46            &  & 2.61       & 84.62      \\
UTK \cite{utk}                      & 5.13            & 63.06            &  & 4.95       & 64.61      \\
CACD \cite{cacd}                    & 4.50            & 4.48             &  & 4.48       & 63.81      \\ \bottomrule
\end{tabular}
}
\label{tab:5}
\end{table}

\begin{table}[]
\centering
\caption{Ablation study on SAW.}
\resizebox{0.55\textwidth}{!}{%
\begin{tabular}{@{}ccccc@{}}
\toprule
Dataset  & $L_{kl}$ & $L_{ce}$ & $L_{kl}+L_{ce}$ & $L_{SAW}$      \\ \midrule
MORPH-II \cite{MORPH} & 2.73              & 2.84                & 2.10          & \textbf{2.01} \\
FG-NET \cite{fgnet}   & 4.29              & 4.17                & 3.59          & \textbf{3.28} \\
UTK \cite{utk}     & 6.77              & 6.60                & 5.23          & \textbf{4.85} \\
CACD \cite{cacd}     & 4.80              & 4.66                & 4.55          & \textbf{4.49} \\ \bottomrule
\end{tabular}
}
\label{tab:6}
\end{table}

\textbf{Stage-wise adaptive weighted loss function.} Table \ref{tab:6} presents the ablation study results using different loss functions on four datasets. The results show that treating age estimation as either a distribution or a classification task leads to varying performance across different datasets, while the combined loss function ($L_{kl}+L_{ce}$) achieves greater advantages. Moreover, the SAW loss function effectively leverages the stage-wise ambiguity of labels by assigning weights to different loss functions based on the severity of label ambiguity in each age stage, thereby achieving better performance.


\begin{table}[!t]
\centering
\caption{Ablation Study on SA-LDL.}
\resizebox{0.70\textwidth}{!}{%
\begin{tabular}{@{}cccccc@{}}
\toprule
SAV                       & SAW                       & MOPRH \cite{MORPH}              & FG-NET \cite{fgnet}             & UTK \cite{utk}                                 & CACD \cite{cacd}          \\ \midrule
                          &                           & 2.21         & 4.63         & 5.13                           & 4.50          \\
\checkmark &                           & 1.89         & 2.66         & \textbf{4.41} & \textbf{4.20} \\
                          & \checkmark & 2.10         & 3.11         & 4.83                           & 4.49          \\
\checkmark & \checkmark & \textbf{1.74} & \textbf{2.15} & 4.43  & \textbf{4.20} \\ \bottomrule
\end{tabular}
}
\label{tab:7}
\end{table}


\textbf{Stage-wise adaptive label distribution learning.} Table \ref{tab:7} presents the ablation results of different design strategies within the proposed SA-LDL method. The experiment compares the effects of using only the stage-wise variance range, only the stage-wise weight range, and the combination of both. The results show that incorporating both stage-wise variance and weight ranges consistently achieves the best performance across all datasets. This indicates that modeling both the diversity and importance of labels within each age stage contributes to more accurate label distribution learning. By comprehensively addressing stage-wise label ambiguity, the combined approach significantly enhances the model’s ability to distinguish similar or overlapping labels within stages and improves overall performance.

\section{Conclusion}
In this paper, we introduce a novel stage-wise adaptive label distribution learning (SA-LDL) algorithm that explicitly models the stage-wise nature of label ambiguity in facial age estimation. SA-LDL incorporates two key components: Stage-wise Adaptive Variance (SAV), which captures the varying uncertainty of labels across age stages, and Stage-wise Adaptive Weighted (SAW) loss function, which dynamically adjusts the importance of loss terms according to stage-specific ambiguity. By jointly leveraging SAV and SAW, the method provides a more precise and robust representation of facial aging patterns. Extensive experiments demonstrate that SA-LDL achieves strong performance, reaching MAE scores of 1.74 on MORPH-II and 2.15 on FG-NET, outperforming existing advanced approaches.

\bibliographystyle{splncs04}
\bibliography{mybibliography}

%





\end{document}